\documentclass{article}

\usepackage{graphicx}
\usepackage{caption}
\usepackage{float}

\PassOptionsToPackage{numbers, compress}{natbib}

\usepackage[dblblindworkshop, final]{neurips_2025}

\newcommand{\datasetname}[1]{\textbf{#1}}
\newcommand*{\datacite}[2]{%
    \datasetname{#1} \cite{#2}%
}

\usepackage[utf8]{inputenc} 
\usepackage[T1]{fontenc}    
\usepackage{hyperref}       
\usepackage{url}            
\usepackage{booktabs}       
\usepackage{amsfonts}       
\usepackage{nicefrac}       
\usepackage{microtype}      
\usepackage{xcolor}   
\usepackage{amsmath}

\begin{document}

\title{LITcoder: A General-Purpose Library for Building and Comparing Encoding Models}

%

\author{%
Taha Binhuraib \\
Georgia Tech \\
\texttt{tbinhuraib3@gatech.edu} \\
\And
Ruimin Gao \\
Georgia Tech \\
\texttt{ruimin.gao@gatech.edu} \\
\And
Anna A. Ivanova \\
Georgia Tech \\
\texttt{a.ivanova@gatech.edu} \\
}

\maketitle

\begin{abstract}
We introduce LITcoder, an open-source library for building and benchmarking neural encoding models. Designed as a flexible backend, LITcoder provides standardized tools for aligning continuous stimuli (e.g., text and speech) with brain data, transforming stimuli into representational features, mapping those features onto brain data, and evaluating the predictive performance of the resulting model on held-out data. The library implements a modular pipeline covering a wide array of methodological design choices, so researchers can easily compose, compare, and extend encoding models without reinventing core infrastructure. Such choices include brain datasets, brain regions, stimulus feature (both neural-net-based and control, such as word rate), downsampling approaches, and many others. In addition, the library provides built-in logging, plotting, and seamless integration with experiment tracking platforms such as Weights \& Biases (W\&B). 
We demonstrate the scalability and versatility of our framework by fitting a range of encoding models to three story listening datasets: LeBel et al. (2023), Narratives, and Little Prince. We also explore the methodological choices critical for building encoding models for continuous fMRI data, illustrating the importance of accounting for all tokens in a TR scan (as opposed to just taking the last one, even when contextualized), incorporating hemodynamic lag effects, using train-test splits that minimize information leakage, and accounting for head motion effects on encoding model predictivity. Overall, LITcoder lowers technical barriers to encoding model implementation, facilitates systematic comparisons across models and datasets, fosters methodological rigor, and accelerates the development of high-quality high-performance predictive models of brain activity.
\newline \newline
\textbf{Project page:} \url{https://litcoder-brain.github.io}
\end{abstract}

\section{Introduction}

Encoding models are a powerful tool for predicting brain responses to external stimuli. Traditionally, encoding models were used to model responses to static images \cite{yamins_performance-optimized_2014, guclu_deep_2015, ratan_murty_computational_2021, kay_identifying_2008, khaligh-razavi_deep_2014}; however, in the last decade their use has been expanded beyond vision to other modalities, such as language \citep{schrimpf_neural_2021}, and from static to continuous stimuli, such as stories \citep{huth_natural_2016, antonello_scaling_2024, caucheteux_brains_2022}. 
These advances make it possible to study brain responses to diverse stimuli in naturalistic settings, track hierarchical processing across cortical networks, and evaluate the extent to which modern AI models align with brain activity \citep{toneva_interpreting_2019, tikochinski_incremental_2025}. However, the rapid growth of encoding-model-based research has resulted in substantial heterogeneity of encoding model design choices made by different research groups, complicating replication and cross-paper comparisons. We provide a methodological framework aimed at resolving these inconsistencies.

\begin{figure}[t]
    \centering
    \includegraphics[width=\linewidth]{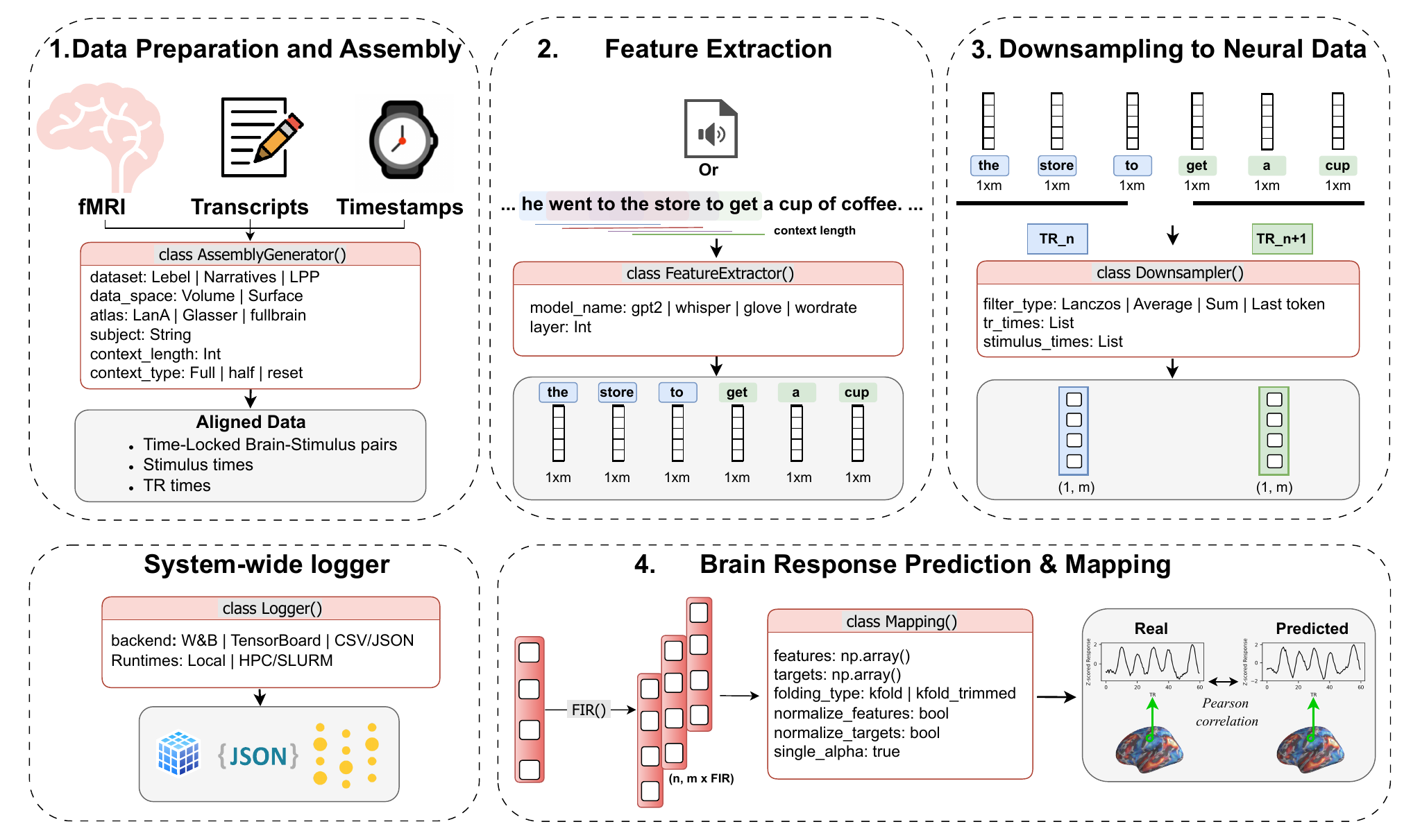}
    \caption{
        \textbf{Overview of the LITcoder library architecture.} 
        The library implements a modular pipeline for constructing and evaluating neural encoding models. 
        \textbf{(1)} Functional MRI data, aligned transcripts, and timestamps are processed through the \texttt{AssemblyGenerator} to produce time-locked brain--stimuli pairs in either volumetric or surface brain spaces. TR=repetition time, a term denoting fMRI data acquisition timepoints.
        \textbf{(2)} Stimuli (text or audio) are transformed into representational features using the \texttt{FeatureExtractor}, which supports a variety of models (e.g., GPT-2, Whisper, GloVe) and non-model-based features (such as word rate). 
        \textbf{(3)} Extracted features are passed to the \texttt{Downsampler}, which implements multiple pooling and filtering strategies (Lanczos, average, sum, last token) to produce TR-matched features. 
        \textbf{(4)} Aligned features are expanded using finite impulse response (FIR) modeling and passed to the \texttt{Mapping} module, which supports flexible cross-validation schemes (e.g., k-fold), feature normalization, and voxel-/vertex-wise prediction using ridge regression. 
        All stages are monitored using the \texttt{Logger}, which interfaces with multiple backends (e.g., Weights \& Biases, TensorBoard, CSV/JSON) and runtime environments (local, HPC, SLURM). 
        }
    \label{fig:figure_1}
\end{figure}

A central challenge for the field is the lack of standardized infrastructure for building and evaluating encoding models. Most works that build encoding models of continuous stimuli only report results from a single dataset \cite{alkhamissi_language_2025}; one reason for this gap is the heterogeneous nature of these datasets' formats, making it difficult to apply the same design choices to all. Furthermore, these works rely on ad hoc pipelines that vary in feature extraction, temporal alignment, cross-validation, and evaluation metrics. Such heterogeneity makes results difficult to compare across papers and limits reproducibility \cite{Palacios2025Executable}. In some cases, methodological choices can introduce systematic biases that inflate predictivity if not carefully controlled. As an example, \citet{hadidi_illusions_2025} demonstrate how temporal autocorrelation can produce spurious model performance, highlighting the need for principled approaches that minimize such confounds. Together, these issues underscore the importance of consistent, transparent pipelines that allow reliable comparison across datasets, models, and experimental paradigms.

To address these limitations, we introduce \textbf{LITcoder}, an open-source library for the systematic evaluation of neural encoding models (Figure \ref{fig:figure_1}). LITcoder provides a modular backend that standardizes core components of the pipeline: feature extraction, temporal alignment and downsampling, hemodynamic lag modeling, voxelwise/vertexwise mapping, and evaluation. The current implementation supports fMRI as the neural data source and language (text and audio) as the input stimulus, but it can be extended to other neural data modalities (e.g., MEG and iEEG) and stimulus types (e.g., videos and audiovisual inputs). The library enforces reproducible, transparent practices while remaining extensible—allowing researchers to plug in new datasets, models, or alignment strategies with minimal overhead. By unifying infrastructure while decoupling it from front-end interfaces, LITcoder makes it vastly easier to compare encoding model performance across datasets, feature spaces, and methodological choices, fostering reproducibility and methodological rigor.

\section{Related Work}

One well-known example of standardized encoding model pipelines is \textbf{Brain-Score}, a powerful framework for benchmarking models against the brain. Brain-Score aggregates multiple datasets and metrics to provide a common interface for ranking candidate neural networks by their ability to predict neural and behavioral responses for vision \citep{schrimpf_brain-score_2018} and language \citep{schrimpf_neural_2021}. 
A key difference between Brain-Score and LITcoder is that the former aims for standardization, whereas we aim for flexibility. When a new neural benchmark and/or model is integrated into Brain-Score, all encoding model design choices are fixed (from stimulus context length to the neural network layer that should be used to map onto neural data). In contrast, LITcoder provides the flexibility to easily experiment with dozens of encoding model choices. 
We therefore view the Brain-Score and the LITcoder frameworks as complementary.



To select encoding model design choices that LITcoder should offer, we take inspiration from existing encoding model frameworks, primarily in the domain of language. We choose four representative papers from different research groups to sample the diversity of design strategies: \citet{schrimpf_neural_2021}, \citet{antonello_scaling_2024}, \citet{oota_joint_2023}, and \citet{caucheteux_brains_2022}. We have identified core design choice decisions made by each of these papers, compiled them together (Table \ref{tab:heterogeneity}), and implemented these options in LITcoder for systematic comparison.

\begin{table}[H]
\centering
\caption{Encoding model design choices across four reference studies.}
\label{tab:heterogeneity}
\resizebox{\textwidth}{!}{%
\begin{tabular}{@{}llllll@{}}
\toprule
\textbf{Study} & \textbf{Downsampling} & \textbf{FIR Delays} & \textbf{Context Updating} & \textbf{Dataset(s)} & \textbf{Projection} \\
\midrule
\citet{schrimpf_neural_2021}     & Last-token & No FIR modeling & Full context & Blank2014\cite{blank_functional_2014}, Pereira2018\cite{pereira_toward_2018} & Volume \\
\citet{antonello_scaling_2024}   & Lanczos    & 4 delays        & Half context & LeBel & Volume \\
\citet{caucheteux_evidence_2023} & Sum/Average & 6 delays        & Full context & Narratives & Surface \\
\citet{oota_joint_2023}        & Lanczos & 8 delays        & Full context & Narratives & Surface \\
\bottomrule
\end{tabular}%
}
\end{table}


\section{The Framework}
\label{section:methods}

LITcoder is designed as a modular backend that provides flexible, interchangeable components for building encoding models while abstracting away implementation details. Each module inherits from a small set of base classes, ensuring that new datasets, stimulus feature extractors, or downsampling procedures can be added with minimal boilerplate. This modularity enables researchers to prototype new ideas rapidly while maintaining a consistent scientific pipeline.

LITcoder proceeds in four steps (Fig.~\ref{fig:figure_1}): 
\textbf{(1) AssemblyGenerator} aligns BIDS fMRI with transcripts in volume or surface space; 
\textbf{(2) FeatureExtractor} turns the aligned stimuli into features (from baselines to GPT-2/Whisper); 
\textbf{(3) Downsampler} aggregates those features to TR resolution; 
\textbf{(4) FIRExpander} adds lags and passes the design matrix to \textbf{Mapping}, which fits ridge models with flexible CV and returns voxel/vertex predictivity. 
All runs are automatically logged via a \texttt{Logger} that interfaces with Weights \& Biases, TensorBoard, or lightweight CSV/JSON backends across local, HPC, and SLURM environments.

\subsection{Data Preparation and Assembly}
This first module takes in functional MRI data (in BIDS format), stimulus transcripts, and timestamps. These inputs are integrated by the \texttt{AssemblyGenerator}, which outputs time-locked brain--stimuli pairs, as well as additional files with stimulus onset times and TR times. 
The module provides flexibility in supporting both volume and surface brain representations, whole-brain and region-based anatomical masks (e.g., from the Glasser atlas \cite{glasser_multi-modal_2016}), and a range of stimulus context lengths when generating stimulus–brain pairs.







Regardless of dataset origin, after assembly all downstream modules receive identically structured inputs. This ensures that observed differences in model performance can be attributed to encoding model design choices rather than idiosyncrasies of data handling.

\paragraph{\emph{Contextual Representation Strategies.}}
LITcoder implements three policies for managing context windows.
\textbf{Full context} uses a fixed-size sliding window and extracts the hidden state of the final token of each word, enabling long-range dependency modeling \citep{schrimpf_neural_2021}. 
\textbf{Half context} advances the window start index by half the window size, reducing computational load while retaining partial overlap \citep{antonello_scaling_2024}. 
\textbf{Reset context} resets the window entirely when the lookback length exceeds the maximum, producing local non-overlapping representations. These strategies provide opportunities to balance comprehensiveness and computational efficiency.

\subsection{Feature Extraction}
\label{sec:features}
A key design goal of LITcoder is to make the extraction of stimulus features both flexible and extensible. Through the \texttt{FeatureExtractor} base class, researchers can draw from a wide range of representational sources: pretrained language models (e.g., GPT-2, Pythia), speech models (e.g., Whisper, HuBERT), static word embeddings (e.g., GloVe, word2vec), and low-level baselines such as word rate or autocorrelation. Implementing a new extractor requires inheriting from the base class and overriding a small number of functions, ensuring that custom models or representations can be seamlessly integrated into the same pipeline. This design separates library infrastructure from scientific experimentation, enabling reproducible extensions without additional effort.

\subsection{Downsampling to Neural Data} \label{sec:downsampling}

A challenge in naturalistic modeling is the temporal mismatch between stimuli and neural data. Language models produce token-level activations ($\sim$3–4 tokens/s), while fMRI samples hemodynamic responses much slower (1.5–2 s TRs). The \texttt{Downsampler} bins token-level activations into TR-sized windows using sum pooling \citep{caucheteux_brains_2022}, average pooling \citep{jain_incorporating_2018}, last-token selection, or Lanczos filtering \citep{huth_continuous_2012}.


\subsection{Brain Response Prediction and Evaluation}
The \texttt{Mapping} module implements voxel- or vertex-wise ridge regression with flexible evaluation protocols. Different folding schemes include standard $k$-fold cross-validation and $k$-fold with boundary trimming to reduce autocorrelation effects (see Section~\ref{sec:autocorr}). Both feature and target normalization are available as options, and models can be run with a single global regularization parameter or voxel-specific parameters. 
LITcoder ensures comparability across datasets and methods. For visualization of voxel-wise and surface-based predictivity maps, we use 
\texttt{nilearn} \citep{abraham_machine_2014}.

\paragraph{\emph{Hemodynamic Response Modeling with Finite Impulse Responses (FIR).}}
\label{sec:fir}
fMRI BOLD signal follows neural activation with a delay and temporal spread. To model this, feature vectors aligned to TRs are expanded into a finite impulse response (FIR) design matrix. For each TR-aligned feature vector $\mathbf{h}_t$, the design matrix includes the current vector and $k$ lagged copies:
\[\mathbf{H}_t = [\mathbf{h}_t, \mathbf{h}_{t-1}, \dots, \mathbf{h}_{t-k}],\]
where $t$ indexes fMRI timepoints in units of TRs and $k$ specifies the number of delays. This procedure allows the regression model to estimate voxel-specific response functions directly from the data rather than assuming a fixed hemodynamic response function. The FIR expansion therefore captures both the delayed onset and the extended temporal profile of the BOLD response, enabling accurate mapping between stimulus features and measured neural activity. In \textbf{LITcoder}, this functionality is encapsulated in a dedicated \texttt{FIRExpander} class, which constructs concatenated, delayed copies of stimulus features with configurable delays.


\begin{figure}[t]
    \centering
    \includegraphics[width=0.9\linewidth]{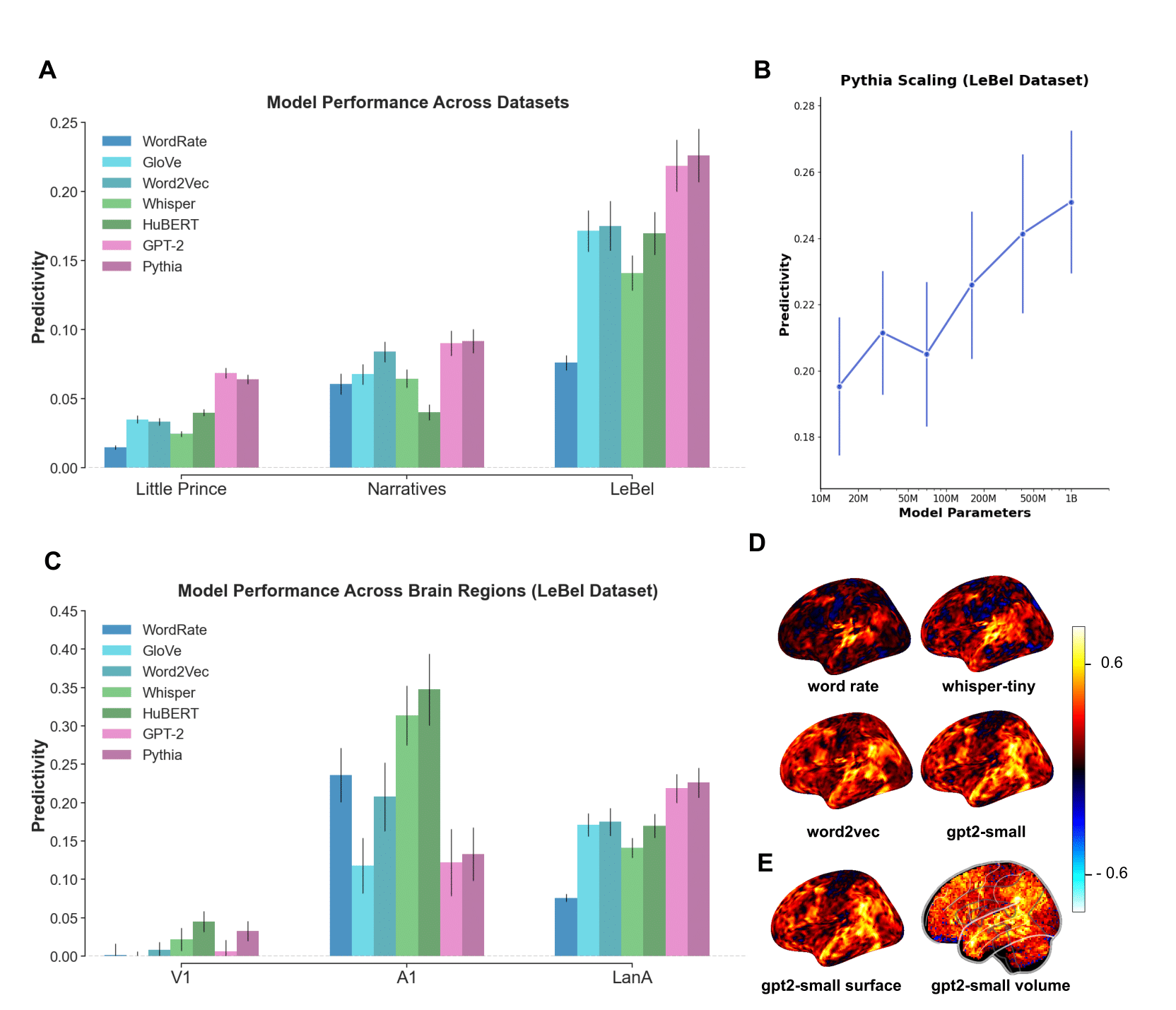}
    \caption{
        \textbf{Encoding model performance across feature families, datasets, and brain regions.} 
        All scores in (A) and (B) are reported as \emph{average voxel-wise correlations within the language network atlas}. 
        \textbf{(A)} Predictivity across three datasets (Little Prince, Narratives, LeBel) for four feature families: simple baseline (word rate), static embeddings (GloVe, word2vec), speech models (Whisper-tiny, HuBERT-base), and language models (GPT-2 small, Pythia-160M). 
        \textbf{(B)} Scaling of the Pythia language model family on the LeBel dataset, showing increased predictivity with model size. 
        \textbf{(C)} Model performance across three cortical regions of interest: V1 (primary visual cortex), 
        A1 (primary auditory cortex), and the language network. For language models, we selected layers that maximized predictivity within each ROI (layer~0 in A1 for both GPT-2 and Pythia; layer~6 in GPT-2 and layer~5 in Pythia for the language network; the same respectively for V1). For speech models,  we used the final encoder layer, following \citet{samara_cortical_2025}.
        \textbf{(D)} Whole-brain voxel-wise predictivity across the entire cortical surface for representative models (word rate, Whisper-tiny, word2vec, GPT2-small) for a single subject from the LeBel dataset. (\textbf{E}) GPT2-small-based encoding model's voxel-wise predictivity correlations in surface (fsaverage5) and volume (MNI) spaces; same subject as (D).
        Here and elsewhere, error bars denote standard error across participants.}
    \label{fig:results_families}
\end{figure}

\section{Experiments}
\label{sec:experiments}
We use LITcoder to run a set of experiments that test the library across feature families, alignment choices, and temporal modeling. Unless noted, \textbf{Predictivity Scores} are reported as \emph{average voxel-wise Pearson correlations within the language network} as defined by the LanA atlas \cite{lipkin_probabilistic_2022}, computed on held-out data following the dataset-specific protocols in Appendix~\ref{app:evaluation-protocols}. These experiments showcase the scientific utility of the pipeline and its extensibility across models and evaluation settings.

\subsection{Datasets}
We evaluated encoding models using three naturalistic fMRI datasets: Narratives \citep{nastase_narratives_2021}, Little Prince \citep{li_petit_2022}, and LeBel \citep{lebel_natural_2023}. All three involve participants listening to extended spoken narratives, providing ecologically valid contexts for studying language comprehension.

The \datacite{Narratives}{nastase_narratives_2021} dataset is one of the largest publicly available collections of auditory story-listening fMRI data, aggregating scans acquired over multiple years and labs. Stimuli comprised 28 naturalistic spoken stories ranging in duration from 3 to 56 minutes, totaling about 5 hours of unique audio material. Following \cite{oota_joint_2023}, we focused on the story “21st year” (56 min), which has been widely adopted in prior work and allows comparability across studies. 18 participants listened to that story and were therefore included in our analysis.

The \datacite{Little Prince}{li_petit_2022} dataset includes fMRI data from participants listening to Antoine de Saint-Exupéry’s book \emph{Little Prince} in English, Chinese, and French, with audiobooks translated and read by professional narrators in each language. We restricted the analyses to the data from 49 English speakers, to maintain comparability with the other datasets. 


The \datacite{LeBel}{lebel_natural_2023} dataset contains naturalistic story listening fMRI data from 8 participants. The stimulus set consisted of 26 autobiographical spoken stories (10-15 minutes each) from The Moth Radio Hour, totaling about 320 minutes, plus one additional 10-minute test story presented in each session (total about 370 minutes per subject).


To ensure comparability, we preprocessed all datasets using \texttt{fMRIPrep} \citep{esteban_fmriprep_2019} with slice-time correction disabled and other parameters at default values (see Appendices~\ref{app:data-processing} and~\ref{app:evaluation-protocols} for details).

\subsection{Experiment 1: LITCoder makes it easy to benchmark feature families across datasets}

LITcoder enables comparisons across datasets, AI models, brain regions, and cortical surfaces. We evaluated four families of stimulus features: simple baselines (word rate), static embeddings (GloVe, word2vec), speech models (Whisper-tiny, HuBERT-base), and language models (GPT-2 small, Pythia) (see Appendix~\ref{sec:models_appendix} for model details). 

First, we show model performance across all feature sets and datasets on the language network voxels, defined with the LanA atlas (Figure \ref{fig:results_families}A). As expected, features from language models (GPT-2 small and Pythia-160m) predict neural responses better than other feature sets. Then, we replicate the findings (e.g.,  \cite{antonello_scaling_2024}) that larger models typically yield better predictivity, using the Pythia models as our feature set and the language-network-constrained neural data from the LeBel dataset (Figure \ref{fig:results_families}B). 


Figure \ref{fig:results_families}C extends the analysis to multiple cortical regions of interest—V1 (primary visual cortex), A1 (primary auditory cortex), and Language network—showing that predictivity is highest in regions associated with auditory and language processing. Figure \ref{fig:results_families}D showcases model performance projected onto the entire cortical surface (fsaverage5, $\sim$22k vertices) for subject UTS03 from the \datacite{LeBel}{lebel_natural_2023} dataset, demonstrating a way to quickly visualize voxelwise predictivities. Finally, Figure \ref{fig:results_families}E shows surface- and volume-based analyses for GPT-2 small, indicating that LITcoder supports both vertexwise and voxelwise encoding models and that these models yield similar results if all other parameters are held constant. 

Together, these analyses illustrate how LITcoder enables systematic comparison of model families and datasets while remaining flexible in spatial scope, from targeted region-of-interest evaluation to full-cortex mapping.

\begin{figure}[t]
    \centering
    \includegraphics[width=\linewidth]{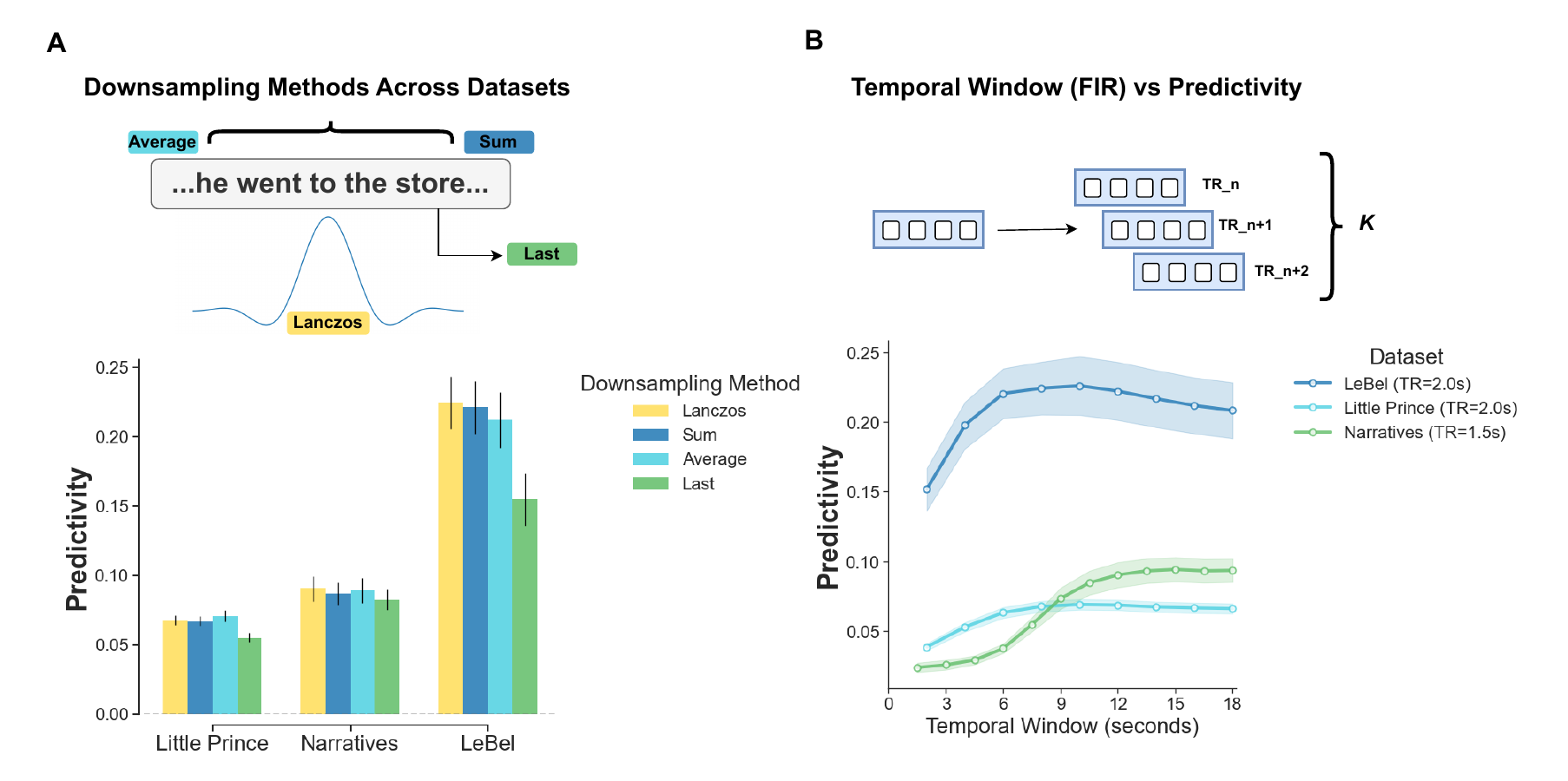}
    \caption{
        \textbf{Evaluating downsampling and temporal modeling choices.}  
        \textbf{(A)} Comparison of token-to-TR downsampling methods. The schematic (top) illustrates four strategies implemented in the \texttt{Downsampler}: average pooling, sum pooling, final-token selection, and Lanczos filtering. The bar plot (bottom) shows predictive performance across datasets (LeBel, Narratives, Little Prince) using GPT-2 small, evaluated at the most predictive layer for each dataset.
        \textbf{(B)} Temporal modeling with finite impulse response (FIR). The schematic (top) illustrates FIR expansion, in which each TR-aligned feature vector is concatenated with $k$ lagged versions. The line plot (bottom) shows the effect of varying the FIR window length $k$ on predictivity for the best-performing GPT-2 small layers in each dataset.
    }
    \label{fig:methods_eval}
\end{figure}

\subsection{Experiment 2: Downsampling and temporal modeling choices affect predictivity}
To demonstrate how LITcoder can be used to iteratively refine methodological design choices, we provide a side-by-side comparison of feature downsampling strategies and varying FIR lengths.

\paragraph{Downsampling.}  
 Figure~\ref{fig:methods_eval}A compares four aggregation strategies with GPT-2 small at the most predictive layer for each dataset. Average and sum pooling perform similarly, with Lanczos filtering matching or slightly exceeding them. Last-token selection produces consistently lower scores, suggesting that ignoring within-TR tokens loses useful information. 

\paragraph{Temporal modeling.}  
 Figure~\ref{fig:methods_eval}B shows the effect of FIR window length. Adding temporal lags improves predictivity until performance stabilizes—around 9--12\,s for LeBel and Little Prince (TR=2.0s), and somewhat shorter for Narratives (TR=1.5s). Extending the window further yields diminishing or negative returns, consistent with overfitting and noise.

Overall, downsampling methods that incorporate all tokens within a TR window generally yield higher predictivity than last-token selection, and finite impulse response (FIR) expansion improves performance up to a dataset-specific plateau.
\label{sec:autocorr}
\subsection{Experiment 3: Mitigating spurious predictivity with contiguous train-test splits}
Time-contiguous stimuli induce strong temporal autocorrelation in both features and BOLD responses, which can inflate cross-validated predictivity if training and test splits are mixed in time. Incorrect folding strategies unfortunately remain a major problem in the literature \cite{hadidi_illusions_2025}. Following \citet{hadidi_illusions_2025}, we implement \emph{Autocorrelation Control Vectors (ACVs)} whose adjacent samples are intentionally similar to quantify the extent to which models can exploit temporal structure alone (see Appendix~\ref{app:temporal-baseline}). 

LITcoder standardizes this through a \texttt{folding\_type} parameter, which supports three strategies: shuffled folds (randomized splits that break temporal contiguity), contiguous folds (non-overlapping time blocks that preserve temporal structure), and contiguous folds with boundary trimming \cite{oota_joint_2023} to reduce leakage from neighboring samples.

\begin{figure}[h]
    \centering
    \includegraphics[width=0.9\linewidth]{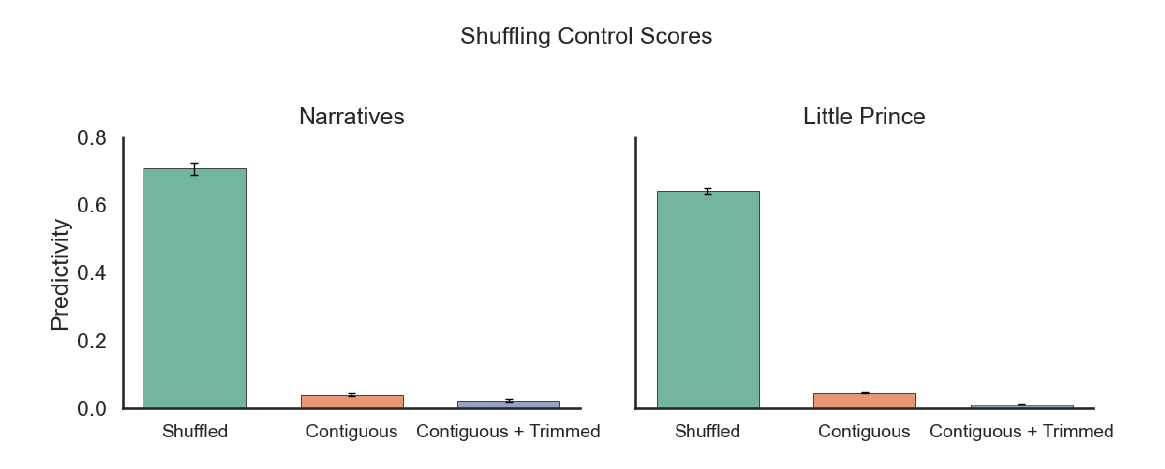}
    \caption{
        \textbf{Effect of train-test splitting choices.} 
        Bars show average voxel-wise correlations across the full cortical surface for \datacite{Narratives}{nastase_narratives_2021} (left) and \datacite{Little Prince}{li_petit_2022} (right) under three cross-validation schemes: 
        \textbf{Shuffled} (randomized folds that ignore temporal order), 
        \textbf{Contiguous} (non-overlapping time blocks), 
         and 
        \textbf{Contiguous + Trimmed} (contiguous folds with boundary trimming). 
    }
    \label{fig:shuffling_controls}
\end{figure}

We apply this analysis to \datacite{Narratives}{nastase_narratives_2021} and \datacite{Little Prince}{li_petit_2022}, the two datasets in our suite that rely on cross-validation rather than a separate held-out story. Results (Figure~\ref{fig:shuffling_controls}) show substantially higher scores under \emph{Shuffled} folds, consistent with temporal leakage; \emph{contiguous} folds reduce this effect, and \emph{trimmed} folds further mitigates boundary bleed. These results highlight the crucial importance of respecting the temporal structure in fMRI data during cross-validation. Shuffled folds artificially inflate model performance by letting temporally adjacent samples leak across train-test boundaries, creating a misleading sense of predictivity. Contiguous folds provide a more conservative and realistic benchmarks, and trimmed folds add further protection against boundary bleed.

\begin{figure}[h]
    \centering
    \includegraphics[width=\linewidth]{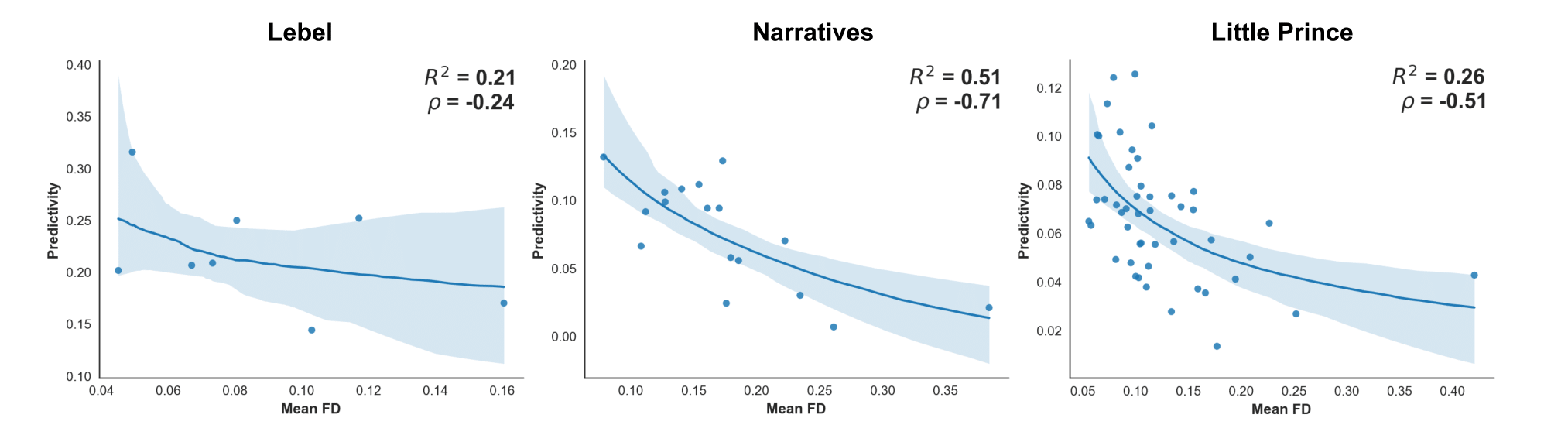}
    \caption{
        \textbf{Head motion vs.\ encoding model predictivity across datasets.}
        Each panel shows subject-level predictivity (language areas' mean voxel correlation from the best-performing layer of GPT-2 small) versus mean framewise displacement (FD) for \datacite{LeBel}{lebel_natural_2023} (left), \datacite{Narratives}{nastase_narratives_2021} (middle), and \datacite{Little Prince}{li_petit_2022} (right).
        Points are individual subjects; the solid curve shows a nonlinear power-law fit, and the shaded region indicates 95\% confidence intervals. $R^2$ values quantify variance explained by the fitted power-law, and $\rho$ denote the Spearman rank correlations between FD and predictivity.
    }
    \label{fig:fd_motion}
\end{figure}

\subsection{Experiment 4: Head motion negatively predicts encoding performance}
\label{sec:motion}

Naturalistic fMRI is susceptible to motion-related artifacts that reduce signal reliability and bias cross-subject comparisons. To quantify the extent to which motion explains variance in our encoding scores, we examined the association between head motion and model predictivity in each dataset. For each participant, we computed \emph{mean framewise displacement (FD)} from the realignment parameters provided by \texttt{fMRIPrep} \citep{esteban_fmriprep_2019}.


We focused on predictivity scores from the best-performing layer of GPT-2 small, evaluated within the  language-network. For each dataset, voxel-wise performance values were averaged to obtain a subject-level predictivity score. We then examined the relationship between mean FD and these subject-level scores by plotting scatterplots with a nonlinear power-law fit of the form $y = a \cdot x^b + c$. 

Across all three datasets, we observe a consistent negative relationship between motion and predictivity: subjects who moved more (higher mean FD) exhibited lower model performance (Figure~\ref{fig:fd_motion}). The effect is most pronounced in datasets with wider motion ranges (Narratives, Little Prince), and is present though slightly reduced in LeBel (fewer subjects, narrower FD range). 

These results align with the known impact of motion on BOLD signal quality \cite{power_spurious_2012}, and they indicate that motion accounts for a meaningful fraction of the between-subject variance in encoding scores. Motion-related variance can confound individual differences, which underscores the need for careful quality control of the neural data that serves as the encoding model target. 


\section{Conclusion}
\textbf{LITcoder} is a flexible open-source library for building and evaluating neural encoding models. By standardizing core components (stimulus-brain alignment, feature extraction, downsampling, mapping, and evaluation) while remaining modular and extensible, LITcoder lowers the barriers to reproducible analysis and makes methodological choices transparent. Experiments across three story listening fMRI datasets showed how decisions about downsampling, FIR window length, cross-validation strategy, and subject motion substantially impact reported predictivity, underscoring the need for principled pipelines. LITcoder therefore provides a toolkit for systematic comparisons across models, datasets, and brain regions and an infrastructure for methodological rigor. 

\section*{Acknowledgments}
We thank N.\ Apurva Ratan Murty, Alish Dipani, Mayukh Deb, Haider Al-Tahan, Badr AlKhamissi, Busra Asan, Jin Li, and other members of the LIT and Murty labs for helpful discussions and feedback.


\bibliographystyle{unsrtnat}


\appendix
\section{Temporal Baseline Features}

\label{app:temporal-baseline}

To quantify the extent to which temporal autocorrelation alone can drive predictivity (independent of stimulus content), we construct a \emph{content-agnostic temporal baseline}. The baseline assigns each stimulus timepoint a feature vector whose similarity depends only on temporal proximity.

\begin{figure}[p]
  \centering
\includegraphics[width=\linewidth]{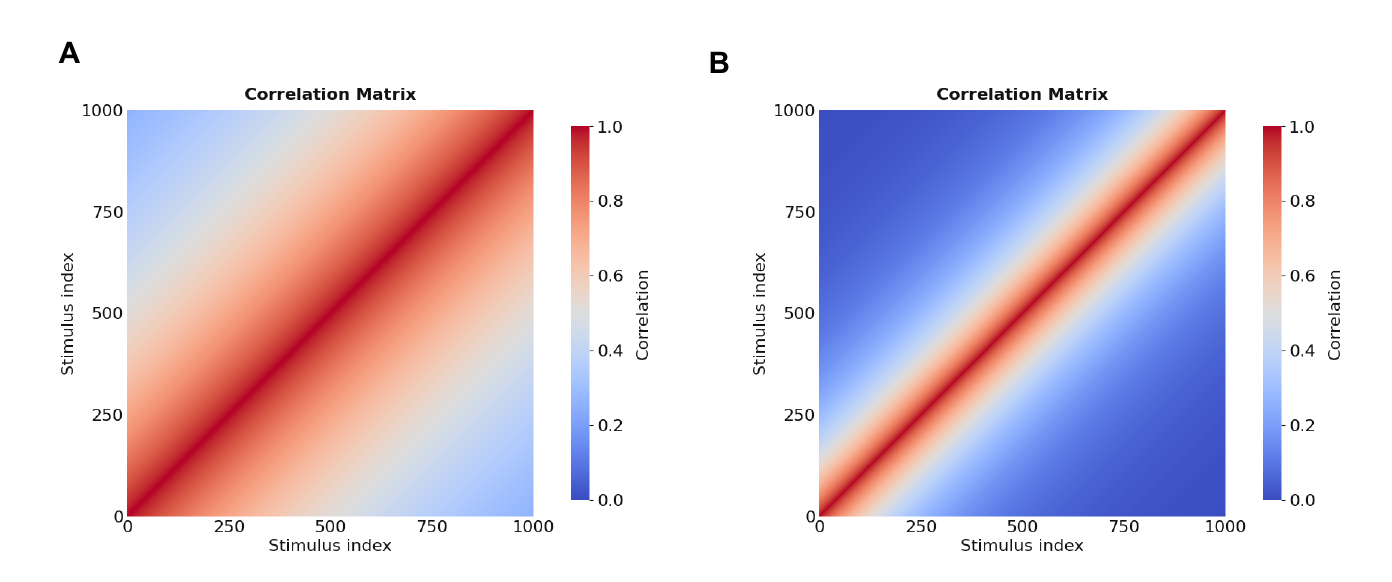}

  \caption{
    \textbf{Temporal autocorrelation kernels used to construct content-agnostic baselines.}
    Each panel shows an example of the exponential kernel
    $A_{ij}=\exp(-|i-j|/\ell)$ that defines similarity between timepoints.
    \textbf{(A)} With a longer correlation length ($\ell=750$), similarity decays slowly,
    producing a broad diagonal band (long-range temporal correlation).
    \textbf{(B)} With a shorter correlation length ($\ell=300$), similarity decays faster,
    yielding a narrower band (short-range correlation).
    These kernels drive the Autocorrelation Control Vectors used in Section~\ref{sec:autocorr}:
    larger $\ell$ sets a stronger/longer-range temporal baseline, smaller $\ell$ a weaker/shorter-range one.
  }
  \label{fig:autocorr_mats}
  \end{figure}

\paragraph{Construction.}
Let $n$ be the number of stimulus timepoints (e.g., tokens or TR-aligned bins). We define an $n \times n$ temporal autocorrelation matrix $A$ with entries
\[
A_{ij} \;=\; \exp\!\Big(-\frac{|i-j|}{\ell}\Big),
\]
where $\ell > 0$ is a correlation length (in stimulus units). This kernel induces exponentially decaying similarity as a function of temporal distance. Figure~\ref{fig:autocorr_mats} illustrates this kernel for two choices of $\ell$.

We compute the truncated SVD of $A$,
\[
A \;=\; U \Sigma V^\top,
\]
and form $d$-dimensional baseline features by taking the top $d$ left singular vectors scaled by the square roots of their singular values:
\[
F \;=\; U_{[:,1:d]} \; \Sigma_{[1:d,1:d]}^{1/2} \;\in \mathbb{R}^{n \times d}.
\]
Each row $F_t$ is the baseline feature for timepoint $t$. By construction, pairwise similarities of $\{F_t\}$ reflect only temporal distance; content is never used.

\paragraph{Hyperparameters.}
The correlation length $\ell$ controls the decay rate of temporal similarity (larger $\ell$ $\Rightarrow$ longer-range correlation). The feature width $d$ controls the rank/complexity of the baseline.

\paragraph{Use in analyses.}
These features serve as \emph{autocorrelation controls}: training an encoding model on $F$ estimates the upper bound of predictivity obtainable from temporal structure alone. Comparing model scores to this baseline helps detect temporal leakage and overoptimistic cross-validation (cf. Section~\ref{sec:autocorr}).

\subsection{Extracting Brain Regions (V1, A1/AC1, and Language Network)}
\label{app:roi-extraction}

We derive cortical ROIs on the \texttt{fsaverage} template from two sources: the HCP--MMP1 (Glasser) atlas for V1 and A1/AC1 (\texttt{lh.HCP-MMP1.annot}, \texttt{rh.HCP-MMP1.annot}), and the LanA probabilistic language atlas for the language network \citep{lipkin_probabilistic_2022}.

\paragraph{V1 and A1/AC1.}
From the Glasser annotations, we select parcels that correspond to V1 and A1/AC1. The resulting parcel indices are converted into boolean masks per hemisphere and then concatenated into whole-cortex masks on \texttt{fsaverage5}.

\paragraph{Language network (LanA).}
LanA provides a per-vertex probability of language selectivity \citep{lipkin_probabilistic_2022}. After transferring the LanA map to \texttt{fsaverage5}, we define the language ROI as the \emph{top 10\% highest-probability vertices} (default setting used throughout). This produces a reproducible, thresholded mask for the language network.

\section{Model Specifications}
\label{sec:models_appendix}

We evaluated both language models (via TransformerLens) and speech models (via Hugging Face). For \textbf{language models}, we relied on the \texttt{HookedTransformer} interface from TransformerLens to expose internal activations. Unless otherwise specified, features were extracted from the \texttt{blocks.\{L\}.hook\_resid\_pre} hook (pre-MLP residual stream) at a given layer $L$. By default, we used the representation of the \emph{last token} for any given input string, though the framework also supports mean-over-tokens pooling. The models included several members of the Pythia family as well as GPT-2 small (Table~\ref{tab:lms_transformerlens}). 

For \textbf{speech models}, we used Hugging Face implementations of OpenAI Whisper (Tiny) and HuBERT (Base). In Whisper, which has 4 encoder and 4 decoder transformer blocks, we extracted features from the final \emph{encoder} layer only. In HuBERT, which has a 12-layer transformer encoder, we extracted features from the final encoder layer. These choices follow prior work using speech representations as predictors of fMRI activity (Table~\ref{tab:speech_models}).

\begin{table}[p]
\centering
\caption{\textbf{Language models used and TransformerLens hook.} 
Unless otherwise noted, we extract features from \texttt{blocks.\{L\}.hook\_resid\_pre} (pre-MLP residual stream) at a specified layer $L$, using the last token within each text segment (configurable to mean-over-tokens).}
\label{tab:lms_transformerlens}
\begin{tabular}{@{}lllll@{}}
\toprule
\textbf{Family} & \textbf{HF ID} & \textbf{Layers} & \textbf{Hook (TransformerLens)} & \textbf{Token pooling} \\
\midrule
Pythia & \texttt{EleutherAI/pythia-14m}  & 6  & \texttt{blocks.\{L\}.hook\_resid\_pre} & last token \\
Pythia & \texttt{EleutherAI/pythia-31m}  & 6  & \texttt{blocks.\{L\}.hook\_resid\_pre} & last token \\
Pythia & \texttt{EleutherAI/pythia-70m}  & 6  & \texttt{blocks.\{L\}.hook\_resid\_pre} & last token \\
Pythia & \texttt{EleutherAI/pythia-160m} & 12 & \texttt{blocks.\{L\}.hook\_resid\_pre} & last token \\
Pythia & \texttt{EleutherAI/pythia-410m} & 24 & \texttt{blocks.\{L\}.hook\_resid\_pre} & last token \\
Pythia & \texttt{EleutherAI/pythia-1b}   & 16 & \texttt{blocks.\{L\}.hook\_resid\_pre} & last token \\
GPT-2  & \texttt{gpt2} (``gpt2-small'')  & 12 & \texttt{blocks.\{L\}.hook\_resid\_pre} & last token \\
\bottomrule
\end{tabular}

\vspace{0.4em}
\small \textit{Notes:} The hook type is configurable in our code (\texttt{hook\_type}), defaulting to \texttt{hook\_resid\_pre}. We also support mean-over-tokens pooling by setting \texttt{last\_token=False}.
\end{table}

\begin{table}[p]
\centering
\caption{\textbf{Speech models used (Hugging Face).} We extract encoder-side representations for mapping to fMRI.}
\label{tab:speech_models}
\begin{tabular}{@{}llll@{}}
\toprule
\textbf{Family} & \textbf{HF ID} & \textbf{Architecture (layers)} & \textbf{Feature layer used} \\
\midrule
Whisper (Tiny) & \texttt{openai/whisper-tiny} & Encoder 4, Decoder 4 & Final \emph{encoder} layer \\
HuBERT (Base)  & \texttt{facebook/hubert-base-ls960} & Encoder 12 & Final encoder layer \\
\bottomrule
\end{tabular}

\vspace{0.4em}

\end{table}

\section{Dataset-Based Evaluation Protocols}
\label{app:evaluation-protocols}
Evaluation procedures were dataset-specific to align with established practices in the literature. For the \datacite{Narratives}{nastase_narratives_2021} dataset, following \cite{oota_joint_2023}, we employed $k$-fold cross-validation with boundary trimming to mitigate autocorrelation effects.

For the \datacite{Little Prince}{li_petit_2022} dataset, we excluded run 1 because the first chapter incorporated visual cues (drawings) to highlight the story’s theme of differences between adults and children \citep{li_petit_2022}. These additional stimuli introduce visual confounds not present in later runs; excluding run 1 therefore ensures that our analyses focus on auditory comprehension and remain comparable across runs. For runs 2-9, we trained encoding models on each run separately using $k$-fold cross-validation with boundary trimming. This approach follows the design philosophy outlined in \cite{tikochinski_incremental_2025}, where neural encoder models are trained individually on each story within the Narratives dataset.

For the larger-scale \datacite{LeBel}{lebel_natural_2023} dataset, we used held-out evaluation: models were trained on a subset of stories and evaluated on an entirely separate story to assess generalization. Following the evaluation protocol in \citet{lebel_natural_2023}, we used the \textit{wheretheressmoke} story as our test story, which was presented once in each of the five scanning sessions. As recommended in the original paper, we averaged responses across these five repetitions to increase signal-to-noise ratio and obtain less biased estimates of model performance.

\section{Data Processing and Selection}
\label{app:data-processing}

We report here the dataset-specific preprocessing steps applied prior to model fitting. These include temporal trimming of runs and motion-based subject exclusion.

\subsection{Temporal Trimming}
Trimming the beginning and end of fMRI runs is standard practice in naturalistic neuroimaging due to several sources of noise and artifact \citep{lebel_natural_2023}. Additionally, many experimental protocols include silent periods or delays that contain no stimulus content relevant to the cognitive processes of interest. 

For the Narratives dataset, we removed the first 14 TRs because they are silent, and the last 9 TRs, following \cite{oota_joint_2023}.

For the Little Prince dataset, each run began with a 4 TR silent period between the trigger and the audiobook onset \cite{li_petit_2022}. Beyond removing these silent periods, we applied additional trimming of the first 5 and last 5 TRs.

For the LeBel dataset, we followed the original protocol, removing 10 TRs from the beginning and end of each story. This eliminated the 10-second silent periods as well as some portion of the story. In addition, following \cite{antonello_scaling_2024} we removed the first 50 TRs from the test story to eliminate long context artifacts. 

\subsection{Subject Exclusion Based on Motion}
Consistent with our finding that head motion is negatively correlated with encoding model performance (Section~\ref{sec:motion}), we excluded participants whose mean framewise displacement (FD) exceeded 0.2\,mm. No participants were excluded from the LeBel dataset. The excluded subjects for the other datasets are listed below.

\begin{table}[h]
\centering
\caption{Subjects excluded based on mean FD $>$ 0.2\,mm.}
\label{tab:fd_exclusions}
\begin{tabular}{@{}lll@{}}
\toprule
\textbf{Dataset} & \textbf{Subjects Excluded} & \textbf{\# Excluded} \\
\midrule
LeBel       & None & 0 \\
Narratives  & Sub-268, Sub-266, Sub-259, Sub-254 & 4 \\
Little Prince & Sub-EN099, Sub-EN075, Sub-EN097, Sub-EN093 & 4 \\
\bottomrule
\end{tabular}
\end{table}

\end{document}